\pdfoutput=1

\documentclass[10pt, a4paper]{article}

\usepackage[]{lrec-coling2024} 

\usepackage{CJKutf8}
\usepackage{graphicx}
\usepackage{booktabs}
\usepackage{tipa}

\usepackage[normalem]{ulem}
\useunder{\uline}{\ul}{}
\usepackage{subfigure}
\usepackage[most]{tcolorbox}
\definecolor{lightgray}{gray}{0.95} 
\usepackage{amssymb}
\usepackage{graphicx}

\newcommand{\revised}[1]{#1}
\newcommand{\daggercmd}[1]{#1$^\dagger$}

\newcommand{\cjk}[1]{\begin{CJK}{UTF8}{gbsn}#1\end{CJK}}
\newcommand{\specialchar}[1]{\textipa{\textvbaraccent{e}}}

\title{Enhancing Taiwanese Hokkien Dual Translation by Exploring \\and Standardizing of Four Writing Systems}

\name{Bo-Han Lu$^\dagger$, Yi-Hsuan Lin$^\dagger$, En-Shiun Annie Lee$^{\ddagger\S}$, Richard Tzong-Han Tsai$^{\dagger\ast\mathcal{C}}$\thanks{$^{\mathcal{C}}$Corresponding author: thtsai@g.ncu.edu.tw}}

\address{$^\dagger$Department of Computer Science and Information Engineering, National Central University, Taiwan \\
$^\ddagger$Department of Computer Science, Faculty of Arts and Science, University of Toronto \\
$^{\S}$Computer Science Program, Faculty of Science, Ontario Tech University \\
$^{\ast}$Center for GIS, Research Center for Humanities and Social Sciences, Academia Sinica, Taiwan \\
\{110522028, 109502543\}@cc.ncu.edu.tw, \\
annie.lee@cs.toronto.edu, thtsai@g.ncu.edu.tw\\}

\abstract{
Machine translation focuses mainly on high-resource languages (HRLs), while low-resource languages (LRLs) like Taiwanese Hokkien are relatively under-explored. The study aims to address this gap by developing a dual translation model between Taiwanese Hokkien and both Traditional Mandarin Chinese and English. We employ a pre-trained LLaMA 2-7B model specialized in Traditional Mandarin Chinese to leverage the orthographic similarities between Taiwanese Hokkien Han and Traditional Mandarin Chinese. Our comprehensive experiments involve translation tasks across various writing systems of Taiwanese Hokkien \revised{as well as} between Taiwanese Hokkien and other HRLs. We find that the use of a limited monolingual corpus still further improves the model's Taiwanese Hokkien capabilities. We then utilize our translation model to standardize all Taiwanese Hokkien writing systems into Hokkien Han, resulting in further performance improvements. Additionally, we introduce an evaluation method incorporating back-translation and GPT-4 to ensure reliable translation quality assessment even for LRLs. The study contributes to narrowing the resource gap for Taiwanese Hokkien and empirically investigates the advantages and limitations of pre-training and fine-tuning based on LLaMA 2.
 \\ \newline  }

\begin{document}

\maketitleabstract
\begin{CJK*}{UTF8}{bkai}
\section{Introduction}
Machine translation (MT), as a crucial subfield of natural language processing (NLP), serves a vital role in overcoming language barriers by translating more texts into the desired language.
However, current MT systems predominantly cater to high-resource languages (HRLs), posing significant challenges for low-resource languages (LRLs).  
Specifically, Taiwanese Hokkien, which is mainly spoken in Taiwan, southern China and a number of countries in Southeast Asia \citep{Ding2016}, faces unique issues owing to historical factors \citep{Ding2016Taiwan} and a persistent absence of standardized writing systems. These factors lead to an extra layer of complexity by introducing inconsistent \revised{corpora, which} hinders the development of NLP research and data-hungry translation models for this language.

In this study, we focus on dual translation between Taiwanese Hokkien and both Mandarin Chinese\footnote{All references to Chinese characters and Mandarin Chinese in this paper refer to the traditional versions.} and English, aiming to bridge the gap between this LRL and other HRLs. Although Taiwanese Hokkien has a significant spoken user base, written forms are less widespread. It is crucial to prioritize NLP research on Taiwanese Hokkien to develop advanced translation models. Taiwanese Hokkien writing systems primarily fall into three categories: Hokkien Han (HAN) using Chinese characters, Tâi-lô (TL) and P\specialchar{}h-ōe-jī (POJ) using Latin script in phonetic forms, and a hybrid system, Hàn-lô (HL). \autoref{tab:hok-example} shows an example sentence represented in these different writing systems.

With the recent advancement of large language models (LLMs) like BLOOM \citep{workshop2023bloom}, ChatGPT and LLaMA \citep{touvron2023llama}, these models have demonstrated their capabilities across various multilingual NLP tasks, including translation tasks \citep{jiaoChatGPTGoodTranslator2023,garciaUnreasonableEffectivenessFewshot2023,yangBigTransAugmentingLarge,xuParadigmShiftMachine2023}. Despite these advancements, state-of-the-art LLMs leave room for improvement in translation tasks, particularly for languages that are considerably removed from HRL \citep{jiaoChatGPTGoodTranslator2023,hendyHowGoodAre2023}.

This study employs a pre-trained LLaMA 2 \citep{touvron2023llama2} model specialized in Mandarin Chinese (ZH), aiming to leverage the orthographic similarities between HAN and ZH to develop a translation model capable of translating between different writing systems of Taiwanese Hokkien as well as between Taiwanese Hokkien and other HRLs like ZH and English. 

\begin{table*}[!htbp]
\centering
\footnotesize
\begin{tabular}{llll|l}
\toprule
\textbf{Language}&\textbf{Writing} & \textbf{Abbreviation}& \textbf{Script} & \textbf{Translated Example} \\ 
\toprule
Hokkien & Tâi-lô & (TL) & Latin & Tsit kuí kang lóng sī pháinn-thinn. \\ 
Hokkien & P\specialchar{}h-ōe-jī &(POJ) & Latin & Chit kúi kang lóng sī phái$^{\text{n}}$-thi$^{\text{n}}$. \\ 
Hokkien & Hàn-lô &(HL) & *Hybrid  & 這幾工\ lóng\ 是歹天。 \\ 
Hokkien & Han &(HAN) & Chinese character & 這幾工攏是歹天。 \\ 
Mandarin & Chinese &(ZH) & Chinese character & 這幾天都是壞天氣。 \\ 
English && (EN) & Latin & \revised{It's been bad weather these days.} \\
\bottomrule
\end{tabular}
\caption{The four different Taiwanese Hokkien writing systems. *Chinese character \& Latin Hybrid}
\label{tab:hok-example}
\end{table*}

\begin{figure*}[htp!]
\begin{center}
\includegraphics[scale=0.102]{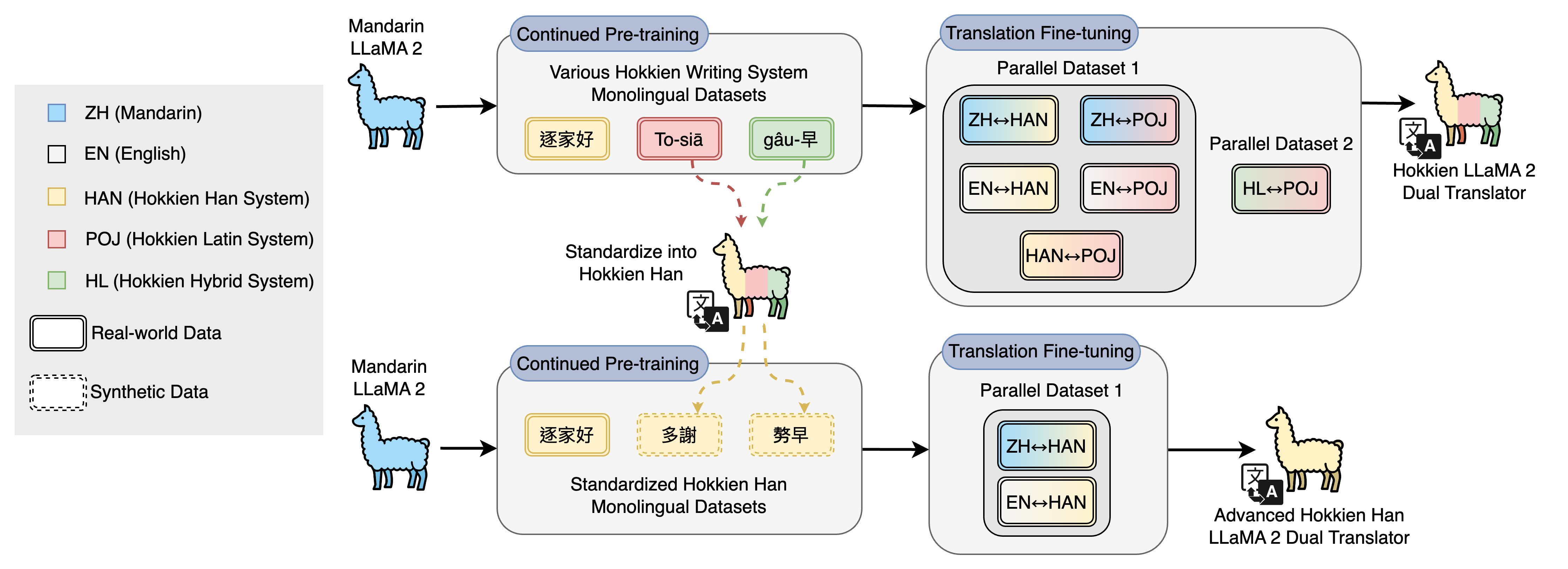} 
\caption{The flowchart of data standardization used to create an advanced Taiwanese Hokkien Han dual translator (HAN-ZH and HAN-EN). The dataset in TL was converted to POJ using an one-to-one mapping rule, allowing for a consistent representation of the Hokkien phonetic sounds.}
\label{fig1_overview}
\end{center}
\end{figure*}
We conduct a comprehensive set of experiments involving translation between the Latin script and Chinese character writing systems of Taiwanese Hokkien as well as translation to and from ZH and English. Our findings indicate that the use of a monolingual corpus covering all Taiwanese Hokkien writing systems positively impacts the model's dual translation performance. Contrary to expectations, extending the model's vocabulary for Taiwanese Hokkien does not yield improvements in these capabilities. We also observe that incorporating parallel datasets involving HRL improves the model's performance, while adding such datasets between the two different Taiwanese Hokkien scripts has detrimental effects.

We further tried to enhance the HAN$\leftrightarrow$ZH and HAN$\leftrightarrow$EN translation by standardizing all Taiwanese Hokkien monolingual corpora into HAN before continued pre-training. The standardization procedure and training flow are illustrated in \autoref{fig1_overview}. Experimental results suggest that this pre-processing step can slightly improve the average translation performance.

For reliable automatic evaluation of translation results, in addition to BLEU score \citep{papineni-etal-2002-bleu} and chrF++ \citep{popovic-2017-chrf} metrics, we modify \citet{kocmiLargeLanguageModels2023}'s evaluation prompt and incorporates a back-translation method \citep{rapp-2009-backtranslation} so that GPT-4 \citep{OpenAI2023GPT4TR} can make reliable evaluations even if the target language is a LRL.

We plan to release the translation model that includes HAN, POJ, ZH and English. We anticipate that this will serve as a reliable translation tool for the public community, and foster the generation of diverse datasets for Taiwanese Hokkien.

To summarize, the major contributions of this work are:
\begin{enumerate}
\item Develop and release the first dual translation model for Taiwanese Hokkien, thereby narrowing the resource gap for this low-resource language.\footnote{\revised{The model and other related resources are available at \href{https://github.com/lbh0830/TW-Hokkien-LLM}{https://github.com/lbh0830/TW-Hokkien-LLM}.}}
\item Empirical evidence to support the enhancement of model performance through monolingual corpora on top of parallel data.
\item Standardized all Taiwanese Hokkien monolingual corpora into HAN prior to continued pre-training, leading to performance enhancements in translations among ZH, English, and HAN.
\item Introduction of back-translation of LRL into HRL for GPT prompt-based evaluation.
\end{enumerate}

\section{Background}
Taiwanese Hokkien, also \revised{known as} Hokkien, Hoklo, Taigi, Southern Min, or Min Nan, is a unique subset of the Southern Min dialects. Sharing a common linguistic heritage with the Fujian dialects, Taiwanese Hokkien has undergone an independent evolution influenced by a number of external factors, including indigenous languages and the colonial legacies of the Dutch and Japanese \citep{liaoFormosaSpeechRecognition2020}. In the following discussion, this dialect will be referred to as ``Hokkien'' for the purpose of simplicity.

Although Hokkien ranks as the second most common language spoken in Taiwan, recent census\footnote{\href{https://www.stat.gov.tw/public/Data/1112144316VT5YTOVB.pdf}{https://www.stat.gov.tw/public/Data/ \\ 1112144316VT5YTOVB.pdf}} suggest a looming crisis due to the decreasing proficiency among younger generations. The challenge is compounded by the fact that most Hokkien speakers only have oral proficiency in Hokkien and lack familiarity with written forms. This dual challenge of decreasing oral proficiency and limited literacy underscores the urgency for targeted research.

\subsection{Writing System Diversity in Hokkien}
The writing systems of Hokkien can be divided into three main groups. The first is Hokkien Han (HAN), which is based on Chinese with additional characters, followed by Latin script systems such as Tâi-lô (TL) and P\specialchar{}h-ōe-jī (POJ). In addition, a hybrid system known as Hàn-lô (HL) combines elements of both systems. Since 2009, an official orthography for HAN has been established and is currently used in educational systems. However, due to its relatively recent standardization, corpus resources of HAN are scarce compared to other systems. On the other hand, POJ, which was introduced by missionaries in the 19th century, has a substantial amount of digitized historical texts and thus provides a rich corpus of Hokkien writings. Moreover, TL, an adaptation of POJ, maintains a systematic correspondence with it. HL, with its mixture of Latin script and Chinese characters, exhibits considerable variance across textual resources due to the lack of a uniform standard for determining the use of Latin script or Chinese characters. Therefore, in this study, the HL corpus is considered for training purposes, but is excluded from translation evaluations.

\subsection{Semantic Divergence of Shared Chinese Characters in HAN and ZH}
Despite the commonality of Chinese characters between HAN and ZH, many homographs differ semantically. For example, the term “手指” translates to “finger” in ZH, while it means “ring” in HAN. Moreover, common HAN terms often correspond to rarely used Chinese characters in ZH, such as ‘覕’ (hide) and ‘\cjk{啉}’ (drink). As a result, training a reliable translation model capable of translating between these two languages under low-resource conditions still remains challenge\revised{s}.

\section{Related Work}

\subsection{Large Language Models in Translation}
LLMs have recently made remarkable progress in translation tasks due to their robust language understanding capabilities from pre-training on massive corpora.  In the field of applying LLMs to translation tasks, \citet{moslemAdaptiveMachineTranslation2023, linFewshotLearningMultilingual2022,zhuMultilingualMachineTranslation2023,zhang2023prompting,vilarPromptingPaLMTranslation2023,garciaUnreasonableEffectivenessFewshot2023} attempted to use in-context learning (ICL) \citep{NEURIPS2020_1457c0d6} to enhance the translation capabilities of LLMs. Their study demonstrated how the pattern of in-context learning, the selection of few-shot sentences, and their quantity could impact the translation results. \citet{zhang2023bayling, yangBigTransAugmentingLarge,liElicitingTranslationAbility2023} tried to enhance the translation abilities of LLMs through instruction-tuning \citep{ouyang2022training} with small amounts of parallel data. \citet{liElicitingTranslationAbility2023} demonstrated $3$ BLEU score on average advancements of multilingual translation when compared to the ICL method.
Some research \citep{hendyHowGoodAre2023,jiaoChatGPTGoodTranslator2023} has indicated that LLMs may have limited translation abilities because their language skills are largely shaped by training in English-centered texts. The translation proficiency of LLMs is often significantly limited when translating languages that are not linguistically close to English \citep{liElicitingTranslationAbility2023}. As a result, \citet{yangBigTransAugmentingLarge} and \citet{liElicitingTranslationAbility2023} have included different translation languages for monolingual training. \citet{xuParadigmShiftMachine2023}'s latest findings also adopt monolingual training before fine-tuning translation tasks. Using medium-sized models with 7B and 13B parameters, they surpassed GPT-3.5 and NLLB-54B \citep{nllbteam2022language} in various translation tasks.

\subsection{Neural Machine Translation in Hokkien}
Due to the scarcity of training data, neural machine translation (NMT) for low-resource languages such as Hokkien faces unique challenges. \citet{9997977} has compiled a dataset for Hokkien speech recognition, which not only contributes to speech-related research but also benefits NMT through the transcribed parallel data. The techniques of transfer learning and cross-lingual models offer possible ways to improve Hokkien NMT systems. \citet{lu-etal-2022-exploring} investigates translation task between ZH, Hokkien code-mixing language and ZH. They apply transfer learning to utilize the knowledge pre-trained on ZH by XLM \citep{lample2019crosslingual}, and develop a method to synthesize a code-mixing translation parallel dataset to achieve better translation results between \revised{the} code-mixing language and ZH. \revised{To the best of our knowledge, we are the first to explore the application of large language models to dual translation for Hokkien, accommodating both its Latin script and Chinese character writing systems.}

\section{Methodology}
\subsection{Corpus Preparation}

\subsubsection{Monolingual Datasets}
As for our continued pre-training data, we have gathered a wide range of linguistic resources from diverse sources that reflect the depth and diversity of the Hokkien language. We have included a comprehensive explanation of the dataset's domain, writing system, and other essential characteristics in \autoref{Tab.CP_data} and \autoref{Fig.distrbution}.

Our corpus primarily comprises web articles in Hokkien collected from diverse internet sources including scripts from Hokkien recitation contests\footnote{\href{https://han-tsi5.knsh.com.tw/Resource.asp?T=MM}{https://han-tsi5.knsh.com.tw/Resource.asp?T=MM}}, lyrics of Hokkien songs shared in Facebook communities\footnote{\href{https://www.facebook.com/groups/922800454445724}{https://www.facebook.com/groups/922800454445724}}, and web-scraped articles covering various domains. The corpus also incorporates religious articles in Hokkien, content from Wikipedia, and Hokkien elementary school textbooks.

Additionally, we include subtitles from Hokkien television programs. Given that these subtitles often lack paragraph breaks and punctuation, we employed GPT-3.5-turbo\footnote{\revised{The term ``GPT-3.5-turbo'' in this paper specifically refers to the version ``gpt-3.5-turbo-0613''.}} to refine the textual structure, rendering it more akin to standard articles.

\begin{figure*}[htp!]
\centering
\subfigure[Domain]{
\label{Fig.sub.domain}
\includegraphics[width=0.45\textwidth]{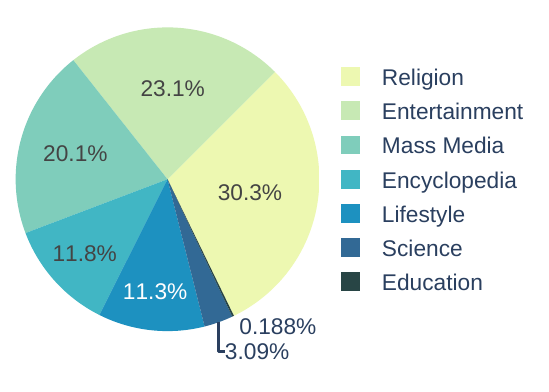}}
\subfigure[Writing system]{
\label{Fig.sub.writing}
\includegraphics[width=0.38\textwidth]{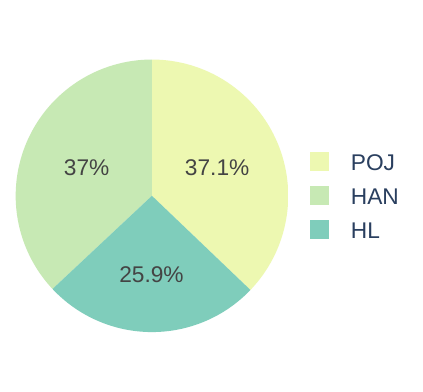}}
\caption{Data distribution of monolingual corpora for continued pre-training
}
\label{Fig.distrbution}
\end{figure*}

\begin{table*}[htp!]
\centering
\begin{tabular}{l|ccc|ccc} 
\hline
\textbf{Data Source} & \multicolumn{3}{c|}{\textbf{Word Count (M)}} & \multicolumn{3}{c}{\textbf{Sentence Count (K)}}  \\ 
                     & HAN    & HL     & POJ                    & HAN     & HL      & POJ                      \\
 \hline
Recitation Contest   & 0.36  & -      & -                      & 38.1   & -       & -                        \\
Hokkien Song Lyrics  & 2.61  & -      & -                      & 349.1  & -       & -                        \\
Web-scraped Articles & 0.37  & 1.18  & 0.48                  & 36     & 110    & 40.6                    \\
Religious Texts      & -      & 2.48  & 3.44                  & -       & 299.5  & 412.9                   \\
Wikipedia Articles   & -      & -      & 1.34                  & -       & -       & 126.2                   \\
Hokkien Textbooks    & 0.02  & -      & -                      & 3.3    & -       & -                        \\
Subtitles            & 1.87  & -      & -                      & 218.3  & -       & -                        \\
\hline
Total                & 5.23  & 3.66  & 5.26                  & 644.8  &  409.5  & 579.7                   \\
\hline
\end{tabular}
\caption{Statistics of monolingual corpora for continued pre-training}
\label{Tab.CP_data}
\end{table*}

\begin{table*}[htp!]
\centering
\begin{tabular}{l|c|c|c} 
\hline
\textbf{Data Source}  & \textbf{Word Count (K)} & \textbf{Pairs} & \textbf{Language}        \\ 
\hline
Dictionary           & 135.4             & 15195          & ZH EN Hokkien (POJ, HAN)  \\
Technical Terms      & 7.9               & 2677           & ZH EN Hokkien (POJ, HAN)  \\
Religious Texts      & 1090.7            & 17872          & Hokkien (POJ, HL)         \\
\hline
\end{tabular}
\caption{Statistics of parallel datasets for fine-tuning}
\label{Tab.FT_data}
\end{table*}

\subsubsection{Parallel Datasets}
For fine-tuning data, we collected the datasets from example sentences in the Hokkien dictionary\footnote{\href{https://sutian.moe.edu.tw/zh-hant/}{https://sutian.moe.edu.tw/zh-hant/}}, religious texts\footnote{\href{http://pojbh.lib.ntnu.edu.tw/script/index.php}{http://pojbh.lib.ntnu.edu.tw/script/index.php}}, and technical terms\footnote{\href{https://stti.moe.edu.tw/?lang=sutgi}{https://stti.moe.edu.tw/?lang=sutgi}}. \revised{Due to the lack of direct Hokkien-to-English parallel dataset, for} those texts that have corresponding parallel sentences in ZH, we used GPT-3.5-turbo to translate them into English. Further details regarding the writing system and other fundamental features for the datasets are provided in \autoref{Tab.FT_data}.

\subsection{Model Training}
\paragraph{Pre-trained Large Language Model}
To leverage the shared Chinese character system between ZH and HAN, we chose TAIDE-7B \revised{academic research model}\footnote{\href{https://taide.tw/}{https://taide.tw/}} as our base model. Enriched with an additional 24k Chinese character tokens and pre-trained on a 1.7B-token Traditional Chinese corpus, TAIDE-7B serves as an extension of LLaMA 2 \citep{touvron2023llama}, enhanced comprehension of Taiwan-specific Traditional Chinese terms.

\paragraph{Vocabulary Extension}
Although the vocabulary of TAIDE-7B model contains a large amount of Chinese characters, it still lacks coverage for Hokkien Latin scripts and some rarely used Chinese characters specific to Hokkien. To address this, we further extend the vocabulary by training a sentence-piece \citep{kudo-richardson-2018-sentencepiece} tokenizer on monolingual Hokkien corpora and merge it back to the original one. Specifically, the vocabulary was extended with an additional 130 Chinese character tokens for HAN and 1876 Latin script tokens for POJ, resulting in a final vocabulary size of 58,505.

\paragraph{Continued Pre-training}
We conducted continued pre-training on monolingual Hokkien corpora across all writing systems.
We followed procedures from Chinese-LLaMA-2 \cite{Chinese-LLaMA-Alpaca} using Low-Rank Adaptation (LoRA) \cite{hu2021lora} with gradient checkpoint to reduce computational cost, and trained 18 epochs to avoid undesirable outputs. Given the rule-based transformation between the POJ and TL systems, we converted all TL in the corpus to POJ using an existing tool\footnote{\href{https://github.com/i3thuan5/KeSi}{https://github.com/i3thuan5/KeSi}}, streamlining the model training across these two Latin script writing systems.

\paragraph{Translation Fine-tuning with Instruction}
In the fine-tuning stage, we modified the LLaMA 2 instruction tuning template, using:

\begin{tcolorbox}[colback=lightgray, sharp corners, colframe=white, boxrule=0pt]
\footnotesize
\texttt{{[TRANS]} \\
X \\
{[/TRANS]} \\
{[\{target\_lang\}]} \\
Y \\
{[/\{target\_lang\}]}}
\end{tcolorbox}

The label [TRANS] denotes translation, X and Y are the source and target sentences, respectively. The label [\{target\_lang\}] indicates the target language for the translation. During fine-tuning, each language pair in the parallel data was fixed to 17,872 instances, including HAN-ZH, HAN-EN, POJ-ZH, POJ-EN, and POJ-HL. Each model was trained for one epoch.

\paragraph{Pre-training Corpus Script-Standardization}
Given the increasing prevalence of HAN in Taiwanese communities in recent years, and aiming to better leverage the orthographic similarities between ZH and HAN, we consequently explored whether standardizing all Hokkien monolingual data into HAN could further improve translation performance in HAN$\leftrightarrow$ZH and HAN$\leftrightarrow$EN directions. To achieve this, we employed our translation model that exhibited the best performance in POJ-HAN translations to standardize all Latin scripts in monolingual Hokkien data into the Chinese character system. Continued pre-training was then carried out on this standardized data.

\subsection{Experimental Settings}
\subsubsection{Two Translation Testing Datasets}
We used iCorpus\footnote{\href{https://github.com/Taiwanese-Corpus/icorpus_ka1_han3-ji7}{https://github.com/Taiwanese-Corpus/icorpus\_ka1\_han3-ji7}}, \revised{a resource originally created by Academia Sinica and subsequently augmented by communities, as our testing data.} It comprises news articles from various domains and includes HAN, POJ and ZH. However under human evaluation, the HAN section of iCorpus contains a considerable number of lexical inaccuracies. To cover this issue in the test set, we first selected terms in HAN that deviate significantly from ZH, using their frequency as a selection criterion to ensure the translation difficulty. Based on this criterion, we sampled the top 100 sentences where HAN appears frequently and manually corrected their lexicons according to official orthography. This resulted in a subset that we named iCorpus-100. Due to the size limitation of iCorpus-100, we integrated an additional data from TAT \citep{liaoFormosaSpeechRecognition2020}, specifically to enhance the evaluation of ZH$\leftrightarrow$HAN and EN$\leftrightarrow$HAN translations. TAT comprises 2,661 parallel sentences sourced from the Taiwanese Across Taiwan speech recognition competitions. Using the same process as the training set, the EN side of the parallel data were generated through translation from ZH using GPT-3.5-turbo. Neither of these test data was used in continued pre-training and fine-tuning stages.

\subsubsection{Evaluation Metrics}\label{subsection_eval}
In order to conduct a comprehensive evaluation of our translation models, we used four different metrics: BLEU score \citep{papineni-etal-2002-bleu}, chrF++ \citep{popovic-2017-chrf}, and two additional GPT-based metrics. Unlike BLEU \revised{score}, chrF++ includes evaluations at the character level. This lead to significant difference between these two metrics when the target language is Latin script, especially POJ. We attribute this to the accent variations in POJ that may alter one or two characters within a word. Such accent-induced variations were not normalized in our corpus. As a result, the model may produce accent inconsistencies with the ground truth, leading to significant penalties in BLEU score, but relatively small drop in chrF++.

Though widely used in translation assessment, BLEU \revised{score} and chrF++ mainly focus on lexicon-level granularity and do not provide a comprehensive evaluation of overall translation quality. In particular, these measures are limited when applied to large language models, which often require more nuanced methods of evaluation \citep{hendyHowGoodAre2023}. \citet{kocmiLargeLanguageModels2023} indicate that GPT-4 holds promise for accurate evaluation of the translation outputs. Therefore, we employ GPT-4\footnote{\revised{The term “GPT-4” in this paper specifically refers to the version “gpt-4-0613”.}} to assess the generated translations, ranging from 0 to 100, by providing both the model's translation and a reference. \revised{However, GPT-4's ability to comprehend Hokkien is restricted. To overcome this constraint, we use the approach inspired by \citet{rapp-2009-backtranslation} to implement back-translation for the model's output when translating from ZH or English into Hokkien. We then compare the back-translated output with the source sentence for evaluation.}
\revised{Note that this approach is not applicable in situations where both the source and target sentences are in Hokkien. For the evaluation of translation scores from GPT-4, we conducted a human qualitative analysis on 20 samples from each of the five score intervals, totaling 100 examples. The qualitative findings based on these analyses are summarized in \autoref{tab:gpt4_score_analysis}. Additionally, the prompt template for GPT-4 evaluations and the examples corresponding to these five score intervals are detailed in Appendix \ref{sec:appendix_gpt4-detail}.}

After evaluating with GPT-4, we calculate the GPT-4 score by taking the average of each translation result. Since translations scoring 80 or above closely approximate the original sentence's meaning, we consider them as correct translations, which are used to compute the translation accuracy, named as GPT-4 accuracy.

\begin{table*}[htp!]
\centering
\scriptsize
\begin{tabular}{>{\centering\hspace{0pt}}m{0.12\linewidth}>{\hspace{0pt}}m{0.88\linewidth}} 
\toprule
\textbf{Score Interval} & \textbf{Quality Descriptors} \\ 
\toprule
80-100 & The translation matches the source in meaning, using identical terms or appropriate synonyms.\par{} The overall grammar is correct, and the meaning flows smoothly. \\ 
\midrule
60-79 & Errors occur with the translation of proper nouns (e.g., names) or misuse of nouns or verbs, causing slight shifts in meaning.\par{} At times, some information may be omitted in the translation or there may be over-translations. \\ 
\midrule
40-59 & Parts of the translation are unrelated to the source text. Misuse of verbs and nouns in inappropriate contexts occurs more frequently. \\ 
\midrule
20-39 & Most of the translation has deviations from the source text, though occasionally it still conveys the basic idea.\par{} Grammar and sentence structures are often incorrect, making the translated content challenging to understand. \\ 
\midrule
0-19 & The translation is almost entirely unrelated to the source text.\par{} Sentences are not fluent, with frequent major grammatical and structural errors. \\
\bottomrule
\end{tabular}
\caption{Qualitative descriptors for translation quality based on GPT-4 score intervals.}
\label{tab:gpt4_score_analysis}
\end{table*}

\begin{table*}[htp!]
\centering
\resizebox{\linewidth}{!}{%
\begin{tabular}{l||cccc||cccc||cccc} 
\toprule
\multicolumn{1}{c||}{\begin{tabular}[c]{@{}c@{}}\textbf{Continued Pre-train Data }\\\textbf{and Vocabulary Extension}\end{tabular}} & \textbf{BLEU} & \textbf{chrF++} & \begin{tabular}[c]{@{}c@{}}\textbf{GPT-4}\\\textbf{Score$^\star$}\end{tabular} & \begin{tabular}[c]{@{}c@{}}\textbf{GPT-4}\\\textbf{Accuracy}\end{tabular} & \textbf{BLEU} & \textbf{chrF++} & \begin{tabular}[c]{@{}c@{}}\textbf{GPT-4}\\\textbf{Score$^\star$}\end{tabular} & \begin{tabular}[c]{@{}c@{}}\textbf{GPT-4}\\\textbf{Accuracy}\end{tabular} & \textbf{BLEU} & \textbf{chrF++} & \begin{tabular}[c]{@{}c@{}}\textbf{GPT-4}\\\textbf{Score$^\star$}\end{tabular} & \begin{tabular}[c]{@{}c@{}}\textbf{GPT-4}\\\textbf{Accuracy}\end{tabular} \\ 
\toprule
 & \multicolumn{4}{c||}{ZH-\textbf{HAN}} & \multicolumn{4}{c||}{\textbf{HAN}-ZH} & \multicolumn{4}{c}{Average} \\ 
\midrule
BASELINE & 26.78 & 27.98 & 32.70 & 2 & 20.89 & 24.44 & 50.20 & 15 & 23.84 & 26.21 & 41.45 & 8.5 \\
NONE & 30.80 & 32.67 & 63.65 & 39 & 29.44 & 32.66 & 67.70 & 46 & 30.12 & 32.67 & 65.68 & 42.5 \\
HAN & 29.15 & 31.20 & 68.40 & 48 & 31.29 & 33.46 & 72.95 & 56 & 30.22 & 32.33 & 70.68 & 52.0 \\
ALL & \uline{31.37} & \uline{32.89} & \uline{72.35} & \uline{55} & 32.60 & 34.91 & \uline{75.35} & \uline{63} & \uline{31.99} & \uline{33.90} & \uline{73.85} & \uline{59.0} \\
ALL \textit{(EXT\_VOCAB)} & 30.90 & 32.43 & 70.30 & 49 & \uline{33.02} & \uline{34.94} & 74.90 & 58 & 31.96 & 33.69 & 72.60 & 53.5 \\ 
\midrule
 & \multicolumn{4}{c||}{EN-\textbf{HAN}} & \multicolumn{4}{c||}{\textbf{HAN}-EN} & \multicolumn{4}{c}{Average} \\ 
\midrule
BASELINE & 6.72 & 13.15 & 48.70 & 16 & 12.32 & 39.88 & 58.25 & 25 & 9.52 & 26.52 & 53.48 & 20.5 \\
NONE & 11.34 & 17.75 & 60.30 & 25 & 14.39 & 41.77 & 58.60 & 21 & 12.87 & 29.76 & 59.45 & 23.0 \\
HAN & \uline{12.83} & \uline{18.72} & 64.50 & 37 & 16.15 & 42.87 & 64.55 & 36 & 14.49 & 30.80 & 64.53 & 36.5 \\
ALL & 12.82 & 18.62 & \uline{65.80} & \uline{40} & \uline{18.45} & \uline{44.92} & \uline{66.75} & \uline{38} & \uline{15.64} & \uline{31.77} & \uline{66.28} & \uline{39.0} \\
ALL \textit{(EXT\_VOCAB)} & 11.93 & 17.94 & 61.30 & 32 & 17.54 & 44.02 & 64.60 & 35 & 14.74 & 30.98 & 62.95 & 33.5 \\ 
\midrule
 & \multicolumn{4}{c||}{ZH-\textbf{POJ}} & \multicolumn{4}{c||}{\textbf{POJ}-ZH} & \multicolumn{4}{c}{Average} \\ 
\midrule
BASELINE & 0.44 & \uline{29.00} & 15.45 & 0 & 12.37 & 16.20 & 12.75 & 1 & 6.41 & 22.60 & 14.10 & 0.5 \\
NONE & 0.38 & 25.43 & 21.20 & 2 & 16.98 & 20.68 & 23.20 & 2 & 8.68 & 23.06 & 22.20 & 2.0 \\
HAN & 0.39 & 25.03 & 19.95 & 3 & 19.48 & 21.73 & 23.95 & 3 & 9.94 & 23.38 & 21.95 & 3.0 \\
ALL & \uline{0.64} & 26.64 & 35.20 & \uline{9} & \uline{30.08} & \uline{31.36} & \uline{43.10} & \uline{12} & \uline{15.36} & 29.00 & \uline{39.15} & \uline{10.5} \\
ALL \textit{(EXT\_VOCAB)} & 0.61 & 28.83 & \uline{38.05} & \uline{9} & 28.15 & 29.85 & 36.00 & 7 & 14.38 & \uline{29.34} & 37.03 & 8.0 \\ 
\midrule
 & \multicolumn{4}{c||}{EN-\textbf{POJ}} & \multicolumn{4}{c||}{\textbf{POJ}-EN} & \multicolumn{4}{c}{Average} \\ 
\midrule
BASELINE & 0.16 & 16.21 & 6.25 & 0 & 1.94 & 21.43 & 4.90 & 0 & 1.05 & 18.82 & 5.58 & 0.0 \\
NONE & 0.19 & 17.75 & 11.05 & 0 & 3.35 & 23.69 & 11.30 & 0 & 1.77 & 20.72 & 11.18 & 0.0 \\
HAN & 0.21 & 17.52 & 8.45 & 0 & 2.92 & 24.36 & 9.25 & 0 & 1.57 & 20.94 & 8.85 & 0.0 \\
ALL & 0.19 & 19.89 & 20.05 & \uline{1} & \uline{7.83} & \uline{30.53} & \uline{24.20} & 3 & \uline{4.01} & \uline{25.21} & 22.13 & \uline{2.0} \\
ALL \textit{(EXT\_VOCAB)} & \uline{0.23} & \uline{20.39} & \uline{25.65} & 0 & 7.06 & 29.39 & 22.95 & \uline{4} & 3.65 & 24.89 & \uline{24.30} & \uline{2.0} \\ 
\midrule
 & \multicolumn{4}{c||}{\textbf{HAN}-\textbf{POJ}} & \multicolumn{4}{c||}{\textbf{POJ}-\textbf{HAN}} & \multicolumn{4}{c}{Average} \\ 
\midrule
BASELINE & 0.91 & \uline{40.03} & - & - & 37.48 & 38.67 & - & - & 19.20 & 39.35 & - & - \\
NONE & 0.55 & 33.95 & - & - & 49.31 & 49.47 & - & - & 24.93 & 41.71 & - & - \\
HAN & 0.74 & 34.44 & - & - & 47.87 & 47.91 & - & - & 24.31 & 41.18 & - & - \\
ALL & \uline{1.01} & 33.19 & - & - & \uline{56.92} & \uline{56.48} & - & - & \uline{28.97} & \uline{44.84} & - & - \\
ALL \textit{(EXT\_VOCAB)} & 0.93 & 36.40 & - & - & 52.13 & 51.93 & - & - & 26.53 & 44.17 & - & - \\
\bottomrule
\end{tabular}
}
\caption{
Continued pre-training ablation study using different input data from various Hokkien writing systems. GPT-based metrics are inapplicable for HAN$\leftrightarrow$POJ evaluation. \uline{underline} = the best results for the respective metric. *We primarily focus on the GPT-4 score.
}
\label{tab:exp1_result}
\end{table*}

\begin{table*}[htp!]
\centering
\resizebox{\linewidth}{!}{%
\begin{tabular}{l||cccc||cccc||cccc} 
\toprule
\multicolumn{1}{c||}{\textbf{Fine-tuning Data}} & \textbf{BLEU} & \textbf{chrF++} & \begin{tabular}[c]{@{}c@{}}\textbf{GPT-4}\\\textbf{Score$^\star$}\end{tabular} & \begin{tabular}[c]{@{}c@{}}\textbf{GPT-4}\\\textbf{Accuracy}\end{tabular} & \textbf{BLEU} & \textbf{chrF++} & \begin{tabular}[c]{@{}c@{}}\textbf{GPT-4}\\\textbf{Score$^\star$}\end{tabular} & \begin{tabular}[c]{@{}c@{}}\textbf{GPT-4}\\\textbf{Accuracy}\end{tabular} & \textbf{BLEU} & \textbf{chrF++} & \begin{tabular}[c]{@{}c@{}}\textbf{GPT-4}\\\textbf{Score$^\star$}\end{tabular} & \begin{tabular}[c]{@{}c@{}}\textbf{GPT-4}\\\textbf{Accuracy}\end{tabular} \\ 
\toprule
 & \multicolumn{4}{c||}{ZH-\textbf{HAN}} & \multicolumn{4}{c||}{\textbf{HAN}-ZH} & \multicolumn{4}{c}{Average} \\ 
\midrule
BASELINE & \daggercmd{18.70} & \daggercmd{21.98} & \daggercmd{65.70} & \daggercmd{48} & \daggercmd{14.88} & \daggercmd{19.15} & \daggercmd{63.85} & \daggercmd{48} & \daggercmd{16.79} & \daggercmd{20.57} & \daggercmd{64.78} & \daggercmd{48.0} \\
ZH, HAN & 27.48 & 29.74 & \uline{75.50} & 61 & 30.56 & 32.74 & 77.90 & 66 & 29.02 & 31.24 & 76.70 & 63.5 \\
ZH, HAN, EN & 26.36 & 28.88 & 74.85 & \uline{64} & 32.41 & 34.46 & \uline{78.90} & \uline{69} & 29.39 & 31.67 & \uline{76.88} & \uline{66.5} \\
ZH, HAN, EN, POJ & 25.16 & 27.86 & 72.60 & 54 & 31.77 & 33.63 & 78.65 & 61 & 28.47 & 30.75 & 75.63 & 57.5 \\
ZH, HAN, EN, POJ, HL & \uline{31.37} & \uline{32.89} & 72.35 & 55 & \uline{32.60} & \uline{34.91} & 75.35 & 63 & \uline{31.99} & \uline{33.90} & 73.85 & 59.0 \\ 
\midrule
 & \multicolumn{4}{c||}{EN-\textbf{HAN}} & \multicolumn{4}{c||}{\textbf{HAN}-EN} & \multicolumn{4}{c}{Average} \\ 
\midrule
BASELINE & \daggercmd{9.49} & \daggercmd{15.04} & \daggercmd{41.35} & \daggercmd{14} & \daggercmd{2.62} & \daggercmd{22.51} & \daggercmd{44.05} & \daggercmd{15} & \daggercmd{6.06} & \daggercmd{18.78} & \daggercmd{42.70} & \daggercmd{14.5} \\
ZH, HAN & \daggercmd{12.23} & \daggercmd{17.89} & \daggercmd{29.00} & \daggercmd{21} & \daggercmd{0.00} & \daggercmd{1.00} & - & - & \daggercmd{6.12} & \daggercmd{9.45} & \daggercmd{29.00} & \daggercmd{21.0} \\
ZH, HAN, EN & \uline{12.83} & 18.46 & \uline{69.05} & \uline{46} & \uline{21.25} & \uline{47.48} & \uline{73.35} & 48 & \uline{17.04} & \uline{32.97} & \uline{71.20} & \uline{47.0} \\
ZH, HAN, EN, POJ & 11.60 & 17.61 & 64.25 & 31 & 20.87 & 46.36 & 72.40 & \uline{52} & 16.24 & 31.99 & 68.33 & 41.5 \\
ZH, HAN, EN, POJ, HL & 12.82 & \uline{18.62} & 65.80 & 40 & 18.45 & 44.92 & 66.75 & 38 & 15.64 & 31.77 & 66.28 & 39.0 \\ 
\midrule
 & \multicolumn{4}{c||}{ZH-\textbf{POJ}} & \multicolumn{4}{c||}{\textbf{POJ}-ZH} & \multicolumn{4}{c}{Average} \\ 
\midrule
BASELINE & \daggercmd{0.08} & \daggercmd{3.36} & \daggercmd{18.30} & \daggercmd{7} & \daggercmd{1.33} & \daggercmd{4.13} & \daggercmd{2.80} & \daggercmd{0} & \daggercmd{0.71} & \daggercmd{3.75} & \daggercmd{10.55} & \daggercmd{3.5} \\
ZH, HAN & \daggercmd{0.00} & \daggercmd{0.47} & - & - & \daggercmd{2.72} & \daggercmd{7.59} & \daggercmd{7.70} & \daggercmd{0} & \daggercmd{1.36} & \daggercmd{4.03} & \daggercmd{7.70} & \daggercmd{0.0} \\
ZH, HAN, EN & \daggercmd{0.01} & \daggercmd{0.83} & - & - & \daggercmd{4.81} & \daggercmd{8.44} & \daggercmd{7.95} & \daggercmd{1} & \daggercmd{2.41} & \daggercmd{4.64} & \daggercmd{7.95} & \daggercmd{1.0} \\
ZH, HAN, EN, POJ & 0.29 & 23.23 & 29.15 & 7 & 22.93 & 25.27 & 36.90 & 10 & 11.61 & 24.25 & 33.03 & 8.5 \\
ZH, HAN, EN, POJ, HL & \uline{0.64} & \uline{26.64} & \uline{35.20} & \uline{9} & \uline{30.08} & \uline{31.36} & \uline{43.10} & \uline{12} & \uline{15.36} & \uline{29.00} & \uline{39.15} & \uline{10.5} \\ 
\midrule
 & \multicolumn{4}{c||}{EN-\textbf{POJ}} & \multicolumn{4}{c||}{\textbf{POJ}-EN} & \multicolumn{4}{c}{Average} \\ 
\midrule
BASELINE & \daggercmd{0.02} & \daggercmd{4.70} & \daggercmd{2.50} & \daggercmd{0} & \daggercmd{0.20} & \daggercmd{6.85} & \daggercmd{1.40} & \daggercmd{0} & \daggercmd{0.11} & \daggercmd{5.78} & \daggercmd{1.95} & \daggercmd{0.0} \\
ZH, HAN & \daggercmd{0.00} & \daggercmd{0.36} & - & - & \daggercmd{0.00} & \daggercmd{0.37} & \daggercmd{6.90} & 0 & \daggercmd{0.00} & \daggercmd{0.37} & \daggercmd{6.90} & 0.0 \\
ZH, HAN, EN & \daggercmd{0.00} & \daggercmd{0.35} & - & - & \daggercmd{1.57} & \daggercmd{19.10} & \daggercmd{5.20} & 0 & \daggercmd{0.79} & \daggercmd{9.73} & \daggercmd{5.20} & 0.0 \\
ZH, HAN, EN, POJ & \uline{0.30} & 19.12 & 17.90 & \uline{1} & 6.86 & 27.86 & \uline{24.95} & 1 & 3.58 & 23.49 & 21.43 & 1.0 \\
ZH, HAN, EN, POJ, HL & 0.19 & \uline{19.89} & \uline{20.05} & \uline{1} & \uline{7.83} & \uline{30.53} & 24.20 & \uline{3} & \uline{4.01} & \uline{25.21} & \uline{22.13} & \uline{2.0} \\ 
\midrule
 & \multicolumn{4}{c||}{\textbf{HAN-POJ}} & \multicolumn{4}{c||}{\textbf{POJ-HAN}} & \multicolumn{4}{c}{Average} \\ 
\midrule
BASELINE & \daggercmd{0.04} & \daggercmd{4.37} & - & - & \daggercmd{1.01} & \daggercmd{4.13} & - & - & \daggercmd{0.53} & \daggercmd{4.25} & - & - \\
ZH, HAN & \daggercmd{0.00} & \daggercmd{0.48} & - & - & \daggercmd{2.96} & \daggercmd{7.92} & - & - & \daggercmd{1.48} & \daggercmd{4.20} & - & - \\
ZH, HAN, EN & \daggercmd{0.02} & \daggercmd{1.10} & - & - & \daggercmd{5.87} & \daggercmd{9.30} & - & - & \daggercmd{2.95} & \daggercmd{5.20} & - & - \\
ZH, HAN, EN, POJ & 0.52 & 27.14 & - & - & 35.98 & 39.32 & - & - & 18.25 & 33.23 & - & - \\
ZH, HAN, EN, POJ, HL & \uline{1.01} & \uline{33.19} & - & - & \uline{56.92} & \uline{56.48} & - & - & \uline{28.97} & \uline{44.84} & - & - \\
\bottomrule
\end{tabular}
}
 
\caption{Fine-tuning ablation study using different input data from various Hokkien writing systems. GPT-based metrics are inapplicable for HAN$\leftrightarrow$POJ evaluation. In the translation directions of ZH-POJ, HAN-EN, and EN-POJ, certain models failed to translate the source sentence into the target language. As a result, the GPT-4 score and GPT-4 accuracy were not computed. \uline{underline} = the best results for the respective metric; $^\dagger$ = the model has not been explicitly trained on that translation direction. *We primarily focus on the GPT-4 score.}
\label{tab:exp2_result}
\end{table*}

\section{Experiment Results and Analysis}
\subsection{Experimental Ablation Studies}

In order to isolate the impact of various data inputs, we conducted ablation studies on continued pre-training of three different Hokkien writing systems, the extension of the input vocabulary with Hokkien Chinese characters and Hokkien Latin systems, and fine-tuning with these three different Hokkien writing systems\footnote{Our focus on Hokkien LRL led us to exclude HRL data from training and evaluation.}.

\subsubsection{Continued Pre-training Corpora and Vocabulary Extension Ablation Studies}
As shown in \autoref{tab:exp1_result}, we took LLaMA 2-7B without any continued pre-training as a baseline. We compared its performance with the TAIDE-7B model continued pre-trained on different monolingual data and evaluated the impact of extending the Hokkien dictionary. Here, “NONE” indicates no continued pre-training on any Hokkien data, “HAN” indicates pre-training solely on Chinese character-based Hokkien data, and “ALL” indicates the inclusion of Latin script (POJ) and hybrid (HL) Hokkien data in addition to “HAN”. \revised{All these models were then fine-tuned using all available parallel data.}

The baseline model, which has no further continued pre-training, performs the worst. In contrast, the TAIDE-7B model improves significantly in all translation directions without relying on Hokkien monolingual data but ZH data. This suggests that using a similar HRL model as a foundational model is beneficial when supplementary monolingual data is not available. Pre-training on HAN data improves the GPT-4 score by $4$ to $6$ points in HAN-related translations. Incorporating all Hokkien data yields the best performance, particularly in POJ-related translations, with a $10$ to $20$ points increase in GPT-4 score. These findings align with previous research \citep{yangBigTransAugmentingLarge,liElicitingTranslationAbility2023,xuParadigmShiftMachine2023}, demonstrating the substantial performance improvement in translation tasks when monolingual data is utilized for languages that the foundational model is not familiar with.

Regarding vocabulary extension, since the added vocabulary primarily consists of Latin scripts from POJ, the model with vocabulary extension exhibits superior performance only when the target language is POJ. For other translation directions, these models exhibit a slight decrement, averaging $3$ points lower on the GPT-4 score compared to models without vocabulary extension. We attribute this to the limited size of the pre-training corpus, which hinders effective tuning of newly added tokens. Consequently, we opted not to extend the vocabulary and suggest future work in collecting a larger POJ corpus for further investigation.

\begin{table*}
\centering
\resizebox{\linewidth}{!}{%
\begin{tabular}{l||cccc||cccc||cccc} 
\toprule
\multicolumn{1}{c||}{\textbf{Continued Pre-train Data}} & \textbf{BLEU} & \textbf{chrF++} & \begin{tabular}[c]{@{}c@{}}\textbf{GPT-4}\\\textbf{Score$^\star$}\end{tabular} & \begin{tabular}[c]{@{}c@{}}\textbf{GPT-4}\\\textbf{Accuracy}\end{tabular} & \textbf{BLEU} & \textbf{chrF++} & \begin{tabular}[c]{@{}c@{}}\textbf{GPT-4}\\\textbf{Score$^\star$}\end{tabular} & \begin{tabular}[c]{@{}c@{}}\textbf{GPT-4}\\\textbf{Accuracy}\end{tabular} & \textbf{BLEU} & \textbf{chrF++} & \begin{tabular}[c]{@{}c@{}}\textbf{GPT-4}\\\textbf{Score$^\star$}\end{tabular} & \begin{tabular}[c]{@{}c@{}}\textbf{GPT-4}\\\textbf{Accuracy}\end{tabular} \\ 
\toprule
 & \multicolumn{4}{c||}{ZH-\textbf{HAN}} & \multicolumn{4}{c||}{\textbf{HAN}-ZH} & \multicolumn{4}{c}{Average} \\ 
\midrule
CP\_HAN (w/o Standardized Data) & 30.63 & 32.16 & 70.58 & 54.38 & 35.20 & 37.55 & 77.96 & 68.92 & 32.92 & 34.86 & 74.27 & 61.65 \\
CP\_ALL & 32.16 & \uline{34.90} & \uline{74.21} & \uline{62.53} & 35.17 & 37.03 & 77.87 & 67.98 & 33.67 & 35.97 & \uline{76.04} & \uline{65.26} \\
CP\_HAN (w/ Standardized Data) & \uline{32.98} & 33.86 & 73.09 & 59.41 & \uline{36.36} & \uline{38.09} & \uline{78.96} & \uline{69.90} & \uline{34.67} & \uline{35.98} & 76.03 & 64.66 \\ 
\midrule
 & \multicolumn{4}{c||}{EN-\textbf{HAN}} & \multicolumn{4}{c||}{\textbf{HAN}-EN} & \multicolumn{4}{c}{Average} \\ 
\midrule
CP\_HAN (w/o Standardized Data) & 19.64 & 23.04 & 63.57 & 41.87 & \uline{24.90} & \uline{46.95} & 67.80 & 50.19 & 22.27 & 35.00 & 65.69 & 46.03 \\
CP\_ALL & 20.05 & 23.40 & 64.52 & 42.81 & 24.42 & 46.57 & 68.14 & 50.00 & 22.24 & 34.99 & 66.33 & 46.41 \\
CP\_HAN (w/ Standardized Data) & \uline{20.99} & \uline{24.31} & \uline{67.32} & \uline{47.05} & 24.36 & 46.90 & \uline{69.12} & \uline{52.31} & \uline{22.68} & \uline{35.61} & \uline{68.22} & \uline{49.68} \\
\bottomrule
\end{tabular}
}
\caption{The translation performance of ZH$\leftrightarrow$HAN and EN$\leftrightarrow$HAN on TAT datasets. \uline{underline} = the best results for the respective metric. *We primarily focus on the GPT-4 score.}
\label{tab:exp3-result}
\end{table*}

\subsubsection{Fine-tuning Datasets Ablation Study}

Given that the model using all Hokkien monolingual data without vocabulary extension performed the best on average across all translation directions, we selected it as the base model for further experiments in the fine-tuning stage. We investigated the impact of incorporating different parallel data during fine-tuning. The parallel data containing HL was only available in the HL$\leftrightarrow$POJ direction. As a baseline, we followed the methodology of \citet{liElicitingTranslationAbility2023}, utilizing in-context learning (ICL) for 8-shot translation without any fine-tuning.

\autoref{tab:exp2_result} indicates that models fine-tuned on parallel datasets significantly outperform few-shot ICL models across all translation directions, particularly in directions involving Latin script Hokkien (POJ), with a GPT-4 score increasing from $17$ to $40$ points. Demonstrate that the benefits of fine-tuning are particularly substantial for writing systems that are not closely related to HRL.

For ZH$\leftrightarrow$HAN and EN$\leftrightarrow$HAN directions, including EN$\leftrightarrow$HAN data enhances translation performance. However, further incorporation of POJ and HL parallel data does not yield additional improvements, suggesting that focusing solely on HRL in the parallel data is more effective for aligning cross-lingual embeddings.

In Hokkien script translation, the model performs the best when all parallel data are included. The inclusion of POJ$\leftrightarrow$HL data dramatically improves translation capabilities in the POJ$\leftrightarrow$HAN. The chrF++ improved by $6.05$ and $17.16$ points for HAN-POJ and POJ-HAN directions, respectively. We hypothesize that HL contains a few Han characters, allowing the model to learn some lexical correlations between different scripts in Hokkien.

\subsection{Pre-training Corpus Script-Standardization}

We evaluated the efficacy of continued pre-training using three different corpora: 
\begin{enumerate}
    \item Monolingual Hokkien data in Chinese characters only.
    \item Monolingual Hokkien data in all writing systems.
    \item All monolingual Hokkien data standardized into Han characters.
\end{enumerate}
These three models were subsequently fine-tuned using only ZH$\leftrightarrow$HAN and EN$\leftrightarrow$HAN parallel data.

\autoref{tab:exp3-result} presents the evaluation results of these three models on TAT. The model pre-trained on standardized monolingual data on par or slightly outperforms the other models on average across both pairs of languages. This suggests that standardizing the writing system into Chinese characters can yield benefits on HAN translation.

\section{Conclusion}
We conducted a thorough examination of the efficiency of large language models in a translation system including Mandarin Chinese, English and two different Hokkien writing systems (HAN and POJ). We employed evaluation metrics including BLEU score, chrF++, and a GPT-4-based scoring method. Our results showed that employing Hokkien similar high-resource language model as a foundational model led to significant performance enhancements. Furthermore, models pre-trained on all available Hokkien data exhibited the highest performance. Extending vocabulary appeared to favor the incorporation of more monolingual data to elicit its efficiency. In the fine-tuning stage, the model demonstrates enhanced HAN translation only when it is supplied with parallel data corresponding to the HRLs that were extensively involved in its prior pre-training. Additionally, we investigated the benefits of standardizing the Hokkien scripts to Chinese characters. Our future work will explore the potential advantage of data augmentation through translations from ZH monolingual corpora into Hokkien Han and assess its impact on translation quality. Moreover, extending this research to include other prevalent spoken languages in Taiwan, like Hakka, can offer a more extensive viewpoint on handling the linguistic variations in Taiwan.

\section{Limitation}

The methodology used in this study is conditional on the fact that one writing system in a low-resource language is similar to a high-resource language, which can be seen as a limitation. In this work, Hokkien Han and Mandarin Chinese share similar writing systems and possess a substantial amount of common vocabulary. This allows for a transfer of knowledge from extensive Mandarin Chinese texts, thereby leveraging the benefits of large language models pre-trained on abundant Mandarin Chinese corpora to achieve an exceptionally efficient translation model.

\section{Ethical Considerations}

One of the major ethical challenges in developing large language models for Hokkien is the limited resources and biased nature of available data. Most of the existing datasets come from news articles that exhibit specific ideological stances, political inclinations, or ethnic biases. The utilization of such skewed data may inadvertently train the model to propagate these biases, thus affecting its fairness. To address ethical concerns, we expanded our dataset to include lyrics, essays, and other neutral literary texts. Our goal was to reduce potential biases and create a more balanced and representative model.

\section{Acknowledgements}
\revised{
We would like to express our sincere gratitude to Dr. Yu-Chun Wang for the generous assistance and insightful discussions on Taiwanese Hokkien. Our thanks also go to Shou-Yi, Hung for his invaluable help in polishing the paper writing. Additionally, we are deeply grateful to all the reviewers for their dedication to improving this research. We are thankful to the Trustworthy AI Dialogue Engine (TAIDE) project for providing the Traditional Chinese academic research foundation LLM, which served as a crucial base for building our model. Special appreciation goes to the National Center for High-performance Computing (NCHC) for providing computational and storage resources. This work was also supported in part by the Co-creation Platform of the Speech-AI Research Center, Industry-Academia Innovation School, NYCU, under the framework of the National Key Fields Industry-University Cooperation and Skilled Personnel Training Act, from the Ministry of Education (MOE), the National Development Fund (NDF), and industry partners in Taiwan. }

\section{Bibliographical References}\label{sec:reference}

\bibliographystyle{lrec-coling2024-natbib}
\bibliography{custom}

\begin{thebibliography}{32}
\expandafter\ifx\csname natexlab\endcsname\relax\def\natexlab#1{#1}\fi

\bibitem[{Brown et~al.(2020)Brown, Mann, Ryder, Subbiah, Kaplan, Dhariwal, Neelakantan, Shyam, Sastry, Askell, Agarwal, Herbert-Voss, Krueger, Henighan, Child, Ramesh, Ziegler, Wu, Winter, Hesse, Chen, Sigler, Litwin, Gray, Chess, Clark, Berner, McCandlish, Radford, Sutskever, and Amodei}]{NEURIPS2020_1457c0d6}
Tom Brown, Benjamin Mann, Nick Ryder, Melanie Subbiah, Jared~D Kaplan, Prafulla Dhariwal, Arvind Neelakantan, Pranav Shyam, Girish Sastry, Amanda Askell, Sandhini Agarwal, Ariel Herbert-Voss, Gretchen Krueger, Tom Henighan, Rewon Child, Aditya Ramesh, Daniel Ziegler, Jeffrey Wu, Clemens Winter, Chris Hesse, Mark Chen, Eric Sigler, Mateusz Litwin, Scott Gray, Benjamin Chess, Jack Clark, Christopher Berner, Sam McCandlish, Alec Radford, Ilya Sutskever, and Dario Amodei. 2020.
\newblock \href {https://proceedings.neurips.cc/paper_files/paper/2020/file/1457c0d6bfcb4967418bfb8ac142f64a-Paper.pdf} {Language models are few-shot learners}.
\newblock In \emph{Advances in Neural Information Processing Systems}, volume~33, pages 1877--1901. {Curran Associates, Inc.}

\bibitem[{Conneau and Lample(2019)}]{lample2019crosslingual}
Alexis Conneau and Guillaume Lample. 2019.
\newblock Cross-lingual language model pretraining.
\newblock \emph{Advances in neural information processing systems}, 32.

\bibitem[{Cui et~al.(2023)Cui, Yang, and Yao}]{Chinese-LLaMA-Alpaca}
Yiming Cui, Ziqing Yang, and Xin Yao. 2023.
\newblock \href {https://arxiv.org/abs/2304.08177} {Efficient and effective text encoding for chinese llama and alpaca}.
\newblock \emph{arXiv preprint arXiv:2304.08177}.

\bibitem[{Ding(2016{\natexlab{a}})}]{Ding2016}
Picus~Sizhi Ding. 2016{\natexlab{a}}.
\newblock \href {https://doi.org/10.1007/978-981-287-594-5_1} {\emph{Introduction}}, pages 1--18. Springer Singapore, Singapore.

\bibitem[{Ding(2016{\natexlab{b}})}]{Ding2016Taiwan}
Picus~Sizhi Ding. 2016{\natexlab{b}}.
\newblock \href {https://doi.org/10.1007/978-981-287-594-5_4} {\emph{Taiwan: The Haven for Southern Min?}}, pages 55--75. Springer Singapore, Singapore.

\bibitem[{Garc{\'i}a et~al.(2023)Garc{\'i}a, Bansal, Cherry, Foster, Krikun, Feng, Johnson, and Firat}]{garciaUnreasonableEffectivenessFewshot2023}
Xavier Garc{\'i}a, Yamini Bansal, Colin Cherry, George~F. Foster, Maxim Krikun, Fan Feng, Melvin Johnson, and Orhan Firat. 2023.
\newblock \href {https://api.semanticscholar.org/CorpusID:256598283} {The unreasonable effectiveness of few-shot learning for machine translation}.
\newblock \emph{ArXiv}, abs/2302.01398.

\bibitem[{Hendy et~al.(2023)Hendy, Abdelrehim, Sharaf, Raunak, Gabr, Matsushita, Kim, Afify, and Awadalla}]{hendyHowGoodAre2023}
Amr Hendy, Mohamed~Gomaa Abdelrehim, Amr Sharaf, Vikas Raunak, Mohamed Gabr, Hitokazu Matsushita, Young~Jin Kim, Mohamed Afify, and Hany~Hassan Awadalla. 2023.
\newblock \href {https://api.semanticscholar.org/CorpusID:257038384} {How good are gpt models at machine translation? a comprehensive evaluation}.
\newblock \emph{ArXiv}, abs/2302.09210.

\bibitem[{Hu et~al.(2021)Hu, Shen, Wallis, Allen-Zhu, Li, Wang, Wang, and Chen}]{hu2021lora}
Edward~J. Hu, Yelong Shen, Phillip Wallis, Zeyuan Allen-Zhu, Yuanzhi Li, Shean Wang, Lu~Wang, and Weizhu Chen. 2021.
\newblock \href {http://arxiv.org/abs/2106.09685} {Lora: Low-rank adaptation of large language models}.

\bibitem[{Jiao et~al.(2023)Jiao, Wang, Huang, Wang, and Tu}]{jiaoChatGPTGoodTranslator2023}
Wenxiang Jiao, Wenxuan Wang, JT~Huang, Xing Wang, and ZP~Tu. 2023.
\newblock Is chatgpt a good translator? yes with gpt-4 as the engine.
\newblock \emph{arXiv preprint arXiv:2301.08745}.

\bibitem[{Kocmi and Federmann(2023)}]{kocmiLargeLanguageModels2023}
Tom Kocmi and Christian Federmann. 2023.
\newblock \href {https://aclanthology.org/2023.eamt-1.19} {Large language models are state-of-the-art evaluators of translation quality}.
\newblock In \emph{Proceedings of the 24th Annual Conference of the European Association for Machine Translation}, pages 193--203, Tampere, Finland. European Association for Machine Translation.

\bibitem[{Kudo and Richardson(2018)}]{kudo-richardson-2018-sentencepiece}
Taku Kudo and John Richardson. 2018.
\newblock \href {https://doi.org/10.18653/v1/D18-2012} {{S}entence{P}iece: A simple and language independent subword tokenizer and detokenizer for neural text processing}.
\newblock In \emph{Proceedings of the 2018 Conference on Empirical Methods in Natural Language Processing: System Demonstrations}, pages 66--71, Brussels, Belgium. Association for Computational Linguistics.

\bibitem[{Li et~al.(2023)Li, Zhou, Huang, Chen, and Chen}]{liElicitingTranslationAbility2023}
Jiahuan Li, Hao Zhou, Shujian Huang, Shan Chen, and Jiajun Chen. 2023.
\newblock \href {https://api.semanticscholar.org/CorpusID:258865882} {Eliciting the translation ability of large language models via multilingual finetuning with translation instructions}.
\newblock \emph{ArXiv}, abs/2305.15083.

\bibitem[{Liao et~al.(2020)Liao, Chang, Tiun, Su, Khoo, Tsay, Tan, Kang, Thiann, Iunn, Yang, and Liang}]{liaoFormosaSpeechRecognition2020}
Yuan-Fu Liao, Chia-Yu Chang, Hak-Khiam Tiun, Huang-Lan Su, Hui-Lu Khoo, Jane~S. Tsay, Le-Kun Tan, Peter Kang, Tsun-guan Thiann, Un-Gian Iunn, Jyh-Her Yang, and Chih-Neng Liang. 2020.
\newblock \href {https://doi.org/10.1109/O-COCOSDA50338.2020.9295019} {Formosa {{Speech Recognition Challenge}} 2020 and {{Taiwanese Across Taiwan Corpus}}}.
\newblock In \emph{2020 23rd {{Conference}} of the {{Oriental COCOSDA International Committee}} for the {{Co-ordination}} and {{Standardisation}} of {{Speech Databases}} and {{Assessment Techniques}} ({{O-COCOSDA}})}, pages 65--70.

\bibitem[{Liao et~al.(2022)Liao, Tsay, Kang, Khoo, Tan, Chang, Iunn, Su, Thiann, Tiun, and Liao}]{9997977}
Yuan-Fu Liao, Jane~S. Tsay, Peter Kang, Hui-Lu Khoo, Le-Kun Tan, Li-Chen Chang, Un-Gian Iunn, Huang-Lan Su, Tsun-Guan Thiann, Hak-Khiam Tiun, and Su-Lian Liao. 2022.
\newblock \href {https://doi.org/10.1109/O-COCOSDA202257103.2022.9997977} {Taiwanese across taiwan corpus and its applications}.
\newblock In \emph{2022 25th Conference of the Oriental COCOSDA International Committee for the Co-ordination and Standardisation of Speech Databases and Assessment Techniques (O-COCOSDA)}, pages 1--5.

\bibitem[{Lin et~al.(2022)Lin, Mihaylov, Artetxe, Wang, Chen, Simig, Ott, Goyal, Bhosale, Du, Pasunuru, Shleifer, Koura, Chaudhary, O'Horo, Wang, Zettlemoyer, Kozareva, Diab, Stoyanov, and Li}]{linFewshotLearningMultilingual2022}
Xi~Victoria Lin, Todor Mihaylov, Mikel Artetxe, Tianlu Wang, Shuohui Chen, Daniel Simig, Myle Ott, Naman Goyal, Shruti Bhosale, Jingfei Du, Ramakanth Pasunuru, Sam Shleifer, Punit~Singh Koura, Vishrav Chaudhary, Brian O'Horo, Jeff Wang, Luke Zettlemoyer, Zornitsa Kozareva, Mona Diab, Veselin Stoyanov, and Xian Li. 2022.
\newblock \href {https://doi.org/10.18653/v1/2022.emnlp-main.616} {Few-shot {{Learning}} with {{Multilingual Generative Language Models}}}.
\newblock In \emph{Proceedings of the 2022 {{Conference}} on {{Empirical Methods}} in {{Natural Language Processing}}}, pages 9019--9052. {Association for Computational Linguistics}.

\bibitem[{Lu et~al.(2022)Lu, Lu, Lu, and Tsai}]{lu-etal-2022-exploring}
Sin-En Lu, Bo-Han Lu, Chao-Yi Lu, and Richard Tzong-Han Tsai. 2022.
\newblock \href {https://doi.org/10.18653/v1/2022.findings-emnlp.469} {Exploring methods for building dialects-{M}andarin code-mixing corpora: A case study in {T}aiwanese hokkien}.
\newblock In \emph{Findings of the Association for Computational Linguistics: EMNLP 2022}, pages 6287--6305, Abu Dhabi, United Arab Emirates. Association for Computational Linguistics.

\bibitem[{Moslem et~al.(2023)Moslem, Haque, Kelleher, and Way}]{moslemAdaptiveMachineTranslation2023}
Yasmin Moslem, Rejwanul Haque, John~D. Kelleher, and Andy Way. 2023.
\newblock \href {https://aclanthology.org/2023.eamt-1.22} {Adaptive {{Machine Translation}} with {{Large Language Models}}}.
\newblock In \emph{Proceedings of the 24th {{Annual Conference}} of the {{European Association}} for {{Machine Translation}}}, pages 227--237. {European Association for Machine Translation}.

\bibitem[{OpenAI(2023)}]{OpenAI2023GPT4TR}
OpenAI. 2023.
\newblock \href {https://api.semanticscholar.org/CorpusID:257532815} {Gpt-4 technical report}.
\newblock \emph{ArXiv}, abs/2303.08774.

\bibitem[{Ouyang et~al.(2022)Ouyang, Wu, Jiang, Almeida, Wainwright, Mishkin, Zhang, Agarwal, Slama, Ray, Schulman, Hilton, Kelton, Miller, Simens, Askell, Welinder, Christiano, Leike, and Lowe}]{ouyang2022training}
Long Ouyang, Jeff Wu, Xu~Jiang, Diogo Almeida, Carroll~L. Wainwright, Pamela Mishkin, Chong Zhang, Sandhini Agarwal, Katarina Slama, Alex Ray, John Schulman, Jacob Hilton, Fraser Kelton, Luke Miller, Maddie Simens, Amanda Askell, Peter Welinder, Paul Christiano, Jan Leike, and Ryan Lowe. 2022.
\newblock \href {http://arxiv.org/abs/2203.02155} {Training language models to follow instructions with human feedback}.

\bibitem[{Papineni et~al.(2002)Papineni, Roukos, Ward, and Zhu}]{papineni-etal-2002-bleu}
Kishore Papineni, Salim Roukos, Todd Ward, and Wei-Jing Zhu. 2002.
\newblock \href {https://doi.org/10.3115/1073083.1073135} {{B}leu: a method for automatic evaluation of machine translation}.
\newblock In \emph{Proceedings of the 40th Annual Meeting of the Association for Computational Linguistics}, pages 311--318, Philadelphia, Pennsylvania, USA. Association for Computational Linguistics.

\bibitem[{Popovi{\'c}(2017)}]{popovic-2017-chrf}
Maja Popovi{\'c}. 2017.
\newblock \href {https://doi.org/10.18653/v1/W17-4770} {chr{F}++: words helping character n-grams}.
\newblock In \emph{Proceedings of the Second Conference on Machine Translation}, pages 612--618, Copenhagen, Denmark. Association for Computational Linguistics.

\bibitem[{Rapp(2009)}]{rapp-2009-backtranslation}
Reinhard Rapp. 2009.
\newblock \href {https://aclanthology.org/P09-2034} {The backtranslation score: Automatic {MT} evalution at the sentence level without reference translations}.
\newblock In \emph{Proceedings of the {ACL}-{IJCNLP} 2009 Conference Short Papers}, pages 133--136, Suntec, Singapore. Association for Computational Linguistics.

\bibitem[{Scao et~al.(2022)Scao, Fan, Akiki, Pavlick, Ilic, Hesslow, Castagn{\'{e}}, Luccioni, Yvon, Gall{\'{e}}, Tow, Rush, Biderman, Webson, Ammanamanchi, Wang, Sagot, Muennighoff, del Moral, Ruwase, Bawden, Bekman, McMillan{-}Major, Beltagy, Nguyen, Saulnier, Tan, Suarez, Sanh, Lauren{\c{c}}on, Jernite, Launay, Mitchell, Raffel, Gokaslan, Simhi, Soroa, Aji, Alfassy, Rogers, Nitzav, Xu, Mou, Emezue, Klamm, Leong, van Strien, Adelani, and et~al.}]{workshop2023bloom}
Teven~Le Scao, Angela Fan, Christopher Akiki, Ellie Pavlick, Suzana Ilic, Daniel Hesslow, Roman Castagn{\'{e}}, Alexandra~Sasha Luccioni, Fran{\c{c}}ois Yvon, Matthias Gall{\'{e}}, Jonathan Tow, Alexander~M. Rush, Stella Biderman, Albert Webson, Pawan~Sasanka Ammanamanchi, Thomas Wang, Beno{\^{\i}}t Sagot, Niklas Muennighoff, Albert~Villanova del Moral, Olatunji Ruwase, Rachel Bawden, Stas Bekman, Angelina McMillan{-}Major, Iz~Beltagy, Huu Nguyen, Lucile Saulnier, Samson Tan, Pedro~Ortiz Suarez, Victor Sanh, Hugo Lauren{\c{c}}on, Yacine Jernite, Julien Launay, Margaret Mitchell, Colin Raffel, Aaron Gokaslan, Adi Simhi, Aitor Soroa, Alham~Fikri Aji, Amit Alfassy, Anna Rogers, Ariel~Kreisberg Nitzav, Canwen Xu, Chenghao Mou, Chris Emezue, Christopher Klamm, Colin Leong, Daniel van Strien, David~Ifeoluwa Adelani, and et~al. 2022.
\newblock \href {https://doi.org/10.48550/arXiv.2211.05100} {{BLOOM:} {A} 176b-parameter open-access multilingual language model}.
\newblock \emph{CoRR}, abs/2211.05100.

\bibitem[{Team et~al.(2022)Team, Costa-jussà, Cross, Çelebi, Elbayad, Heafield, Heffernan, Kalbassi, Lam, Licht, Maillard, Sun, Wang, Wenzek, Youngblood, Akula, Barrault, Gonzalez, Hansanti, Hoffman, Jarrett, Sadagopan, Rowe, Spruit, Tran, Andrews, Ayan, Bhosale, Edunov, Fan, Gao, Goswami, Guzmán, Koehn, Mourachko, Ropers, Saleem, Schwenk, and Wang}]{nllbteam2022language}
NLLB Team, Marta~R. Costa-jussà, James Cross, Onur Çelebi, Maha Elbayad, Kenneth Heafield, Kevin Heffernan, Elahe Kalbassi, Janice Lam, Daniel Licht, Jean Maillard, Anna Sun, Skyler Wang, Guillaume Wenzek, Al~Youngblood, Bapi Akula, Loic Barrault, Gabriel~Mejia Gonzalez, Prangthip Hansanti, John Hoffman, Semarley Jarrett, Kaushik~Ram Sadagopan, Dirk Rowe, Shannon Spruit, Chau Tran, Pierre Andrews, Necip~Fazil Ayan, Shruti Bhosale, Sergey Edunov, Angela Fan, Cynthia Gao, Vedanuj Goswami, Francisco Guzmán, Philipp Koehn, Alexandre Mourachko, Christophe Ropers, Safiyyah Saleem, Holger Schwenk, and Jeff Wang. 2022.
\newblock \href {http://arxiv.org/abs/2207.04672} {No language left behind: Scaling human-centered machine translation}.

\bibitem[{Touvron et~al.(2023{\natexlab{a}})Touvron, Lavril, Izacard, Martinet, Lachaux, Lacroix, Rozière, Goyal, Hambro, Azhar, Rodriguez, Joulin, Grave, and Lample}]{touvron2023llama}
Hugo Touvron, Thibaut Lavril, Gautier Izacard, Xavier Martinet, Marie-Anne Lachaux, Timothée Lacroix, Baptiste Rozière, Naman Goyal, Eric Hambro, Faisal Azhar, Aurelien Rodriguez, Armand Joulin, Edouard Grave, and Guillaume Lample. 2023{\natexlab{a}}.
\newblock \href {http://arxiv.org/abs/2302.13971} {Llama: Open and efficient foundation language models}.

\bibitem[{Touvron et~al.(2023{\natexlab{b}})Touvron, Martin, Stone, Albert, Almahairi, Babaei, Bashlykov, Batra, Bhargava, Bhosale et~al.}]{touvron2023llama2}
Hugo Touvron, Louis Martin, Kevin Stone, Peter Albert, Amjad Almahairi, Yasmine Babaei, Nikolay Bashlykov, Soumya Batra, Prajjwal Bhargava, Shruti Bhosale, et~al. 2023{\natexlab{b}}.
\newblock Llama 2: Open foundation and fine-tuned chat models.
\newblock \emph{arXiv preprint arXiv:2307.09288}.

\bibitem[{Vilar et~al.(2023)Vilar, Freitag, Cherry, Luo, Ratnakar, and Foster}]{vilarPromptingPaLMTranslation2023}
David Vilar, Markus Freitag, Colin Cherry, Jiaming Luo, Viresh Ratnakar, and George Foster. 2023.
\newblock \href {https://doi.org/10.18653/v1/2023.acl-long.859} {Prompting {{PaLM}} for {{Translation}}: {{Assessing Strategies}} and {{Performance}}}.
\newblock In \emph{Proceedings of the 61st {{Annual Meeting}} of the {{Association}} for {{Computational Linguistics}} ({{Volume}} 1: {{Long Papers}})}, pages 15406--15427. {Association for Computational Linguistics}.

\bibitem[{Xu et~al.(2023)Xu, Kim, Sharaf, and Awadalla}]{xuParadigmShiftMachine2023}
Haoran Xu, Young~Jin Kim, Amr Sharaf, and Hany~Hassan Awadalla. 2023.
\newblock A paradigm shift in machine translation: Boosting translation performance of large language models.
\newblock \emph{arXiv preprint arXiv:2309.11674}.

\bibitem[{Yang et~al.(2023)Yang, Li, Zhang, and Zong}]{yangBigTransAugmentingLarge}
Wen Yang, Chong Li, Jiajun Zhang, and Chengqing Zong. 2023.
\newblock Bigtrans: Augmenting large language models with multilingual translation capability over 100 languages.
\newblock \emph{arXiv preprint arXiv:2305.18098}.

\bibitem[{Zhang et~al.(2023{\natexlab{a}})Zhang, Haddow, and Birch}]{zhang2023prompting}
Biao Zhang, Barry Haddow, and Alexandra Birch. 2023{\natexlab{a}}.
\newblock \href {http://arxiv.org/abs/2301.07069} {Prompting large language model for machine translation: A case study}.

\bibitem[{Zhang et~al.(2023{\natexlab{b}})Zhang, Fang, Zhang, Ma, Zhou, Huang, Bu, Gui, Chen, Chen et~al.}]{zhang2023bayling}
Shaolei Zhang, Qingkai Fang, Zhuocheng Zhang, Zhengrui Ma, Yan Zhou, Langlin Huang, Mengyu Bu, Shangtong Gui, Yunji Chen, Xilin Chen, et~al. 2023{\natexlab{b}}.
\newblock Bayling: Bridging cross-lingual alignment and instruction following through interactive translation for large language models.
\newblock \emph{arXiv preprint arXiv:2306.10968}.

\bibitem[{Zhu et~al.(2023)Zhu, Liu, Dong, Xu, Kong, Chen, Li, and Huang}]{zhuMultilingualMachineTranslation2023}
Wenhao Zhu, Hongyi Liu, Qingxiu Dong, Jingjing Xu, Lingpeng Kong, Jiajun Chen, Lei Li, and Shujian Huang. 2023.
\newblock Multilingual machine translation with large language models: Empirical results and analysis.
\newblock \emph{arXiv preprint arXiv:2304.04675}.

\end{thebibliography}



\appendix

\section{Details of the GPT-4 evaluation methodology}
\label{sec:appendix_gpt4-detail}
\subsection{Prompt Template for Evaluations}\label{sec:appendix_prompt template_evaluation}

When the target language is in English or Mandarin Chinese:

\begin{tcolorbox}[colback=lightgray, sharp corners, colframe=white, boxrule=0pt]
\texttt{Score the following translation result based on its preservation of the reference sentence without explanation:}\\

\texttt{Reference sentence: "\{reference\_sentence\}"}\\

\texttt{Translation result: "\{forward\_sentence\}"}\\

\texttt{The score should be ranging from 0 to 100, where a score of zero means "no meaning preserved" and score of one hundred means "perfect meaning and grammar".}
\end{tcolorbox}

When the source language is in English or Mandarin Chinese and the target language is in Hokkien:

\begin{tcolorbox}[colback=lightgray, sharp corners, colframe=white, boxrule=0pt]
\texttt{Here are two sentences: Sentence 1 is the original, and Sentence 2 has been translated into another language and then back into the original language. Please score Sentence 2 based on its preservation of the original meaning without explanation:} \\

\texttt{Sentence 1: "\{reference\_sentence\}"} \\

\texttt{Sentence 2: "\{backward\_sentence\}"} \\

\texttt{The score should be ranging from 0 to 100, where a score of zero means "no meaning preserved" and score of one hundred means "perfect meaning and grammar".}
\end{tcolorbox}

\subsection{Examples of GPT-4 Scoring on Translations}\label{sec:gpt4_score_desc}

\revised{To provide a more comprehensive understanding of the GPT-4 scoring standards, we have included examples of translations from Hokkien Han to Mandarin Chinese and English in \autoref{tab:gpt4_score_han_zh} and \autoref{tab:gpt4_score_han_en}, respectively. These examples demonstrate the varied scoring outcomes provided by GPT-4, alongside the translation qualities that correspond to different scores.}

\begin{table*}
\centering
\resizebox{\linewidth}{!}{%
\begin{tabular}{l||c} 
\toprule
\multicolumn{2}{l}{\begin{tabular}[c]{@{}l@{}}\textbf{Hokkien Han Source:} 伊有厝宅貸款壓力，猶有囡仔欲飼，未來欲按怎？\\\textbf{Mandarin Chinese Reference: }他有房屋貸款壓力，還有小孩要養，未來該怎麼辦？\\\textbf{English: }He has the pressure of a mortgage and children to support, what should he do in the future?\end{tabular}} \\ 
\toprule
\textbf{Model Translation Results} & \begin{tabular}[c]{@{}c@{}}\textbf{GPT-4}\\\textbf{Score}\end{tabular} \\ 
\toprule
\begin{tabular}[c]{@{}l@{}}他有房屋貸款壓力，還有小孩要養，未來要怎麼辦？\\\textit{He has the pressure of a mortgage and children to support, what will he do in the future?}\end{tabular} & 100 \\ 
\midrule
\begin{tabular}[c]{@{}l@{}}他家中存在房屋賭債壓力，還有小孩要養育，未來如何是好？\\\textit{His family is under pressure to gamble on their home, and he has a child to raise, so what's the future?}\end{tabular} & 70 \\ 
\midrule
\begin{tabular}[c]{@{}l@{}}他有財力，又有小孩要撫養，以後該怎麼辦？\\\textit{He has money and a child to raise. What should he do in the future?}\end{tabular} & 50 \\ 
\midrule
\begin{tabular}[c]{@{}l@{}}他有房子，又有小孩要撫養，怎麼會貧窮？\\\textit{He owns a house and has children to raise. How can he be poor?}\end{tabular} & 30 \\ 
\midrule
\begin{tabular}[c]{@{}l@{}}他有文學才能，還年輕得想懷孕生子，怎麼會？\\\textit{He's got literary talent, and he's young enough to want to get pregnant and have a baby. How could he?}\end{tabular} & 0 \\
\bottomrule
\end{tabular}
}
\caption{Examples of model translation results from Hokkien Han to Mandarin Chinese, featuring different GPT-4 scores and their corresponding outputs.}
\label{tab:gpt4_score_han_zh}
\end{table*}

\begin{table*}[htp!]
\centering
\resizebox{\linewidth}{!}{%
\begin{tabular}{>{\hspace{0pt}}m{0.9\linewidth}||>{\centering\arraybackslash\hspace{0pt}}m{0.065\linewidth}} 
\toprule
\multicolumn{2}{>{\hspace{0pt}}m{1\linewidth}}{\textbf{Hokkien Han Source:} 準備離婚的翁仔姐仔共同攑槌，對準結婚手指殘殘摃落，佇觀眾的噗仔聲中，結婚手指當場歪斜走形。\par{}\textbf{English Reference: }The couple who were preparing for divorce both lifted their hammers and struck hard at their wedding rings. Amidst the applause of the audience, the wedding rings twisted and deformed on the spot.} \\ 
\toprule
\textbf{Model Translation Results} & \textbf{GPT-4}\par{}\textbf{Score} \\ 
\toprule
The couple prepared to divorce and took the hammer together. They hit each other's wedding rings with force, causing them to deform in front of the audience's applause. & 90 \\ 
\cmidrule(r){1-2}
The doll sisters prepared to divorce, and together held the hammer. They hit their wedding ring hard, causing it to bend in front of the audience's applause. & 60 \\ 
\cmidrule(lr){1-2}
The bride and groom held the hammer together to sign their divorce agreement. They both hit each other's ring finger hard, causing it to twist in front of the audience's applause. & 50 \\ 
\cmidrule(lr){1-2}
The doll sisters who are preparing for divorce hold a hammer together, and the ring finger is broken in front of the audience's applause. On stage, they immediately ran away with their hands crossed. & 20 \\ 
\cmidrule(r){1-2}
The bride and groom held the hammer together to hit the nails, but they accidentally hit each other's fingers. In front of the audience, their hands were twisted in a strange way. & 10 \\
\bottomrule
\end{tabular}
}
\caption{Examples of model translation results from Hokkien Han to English, featuring different GPT-4 scores and their corresponding outputs.}
\label{tab:gpt4_score_han_en}
\end{table*}

\section{JSD for Corpora}

\subsection{JSD of Continued Pre-training Monolingual Corpora }\label{sec:appendix_cp_data_jsd}

\begin{figure}[htp!]
\begin{center}
\includegraphics[scale=0.32]{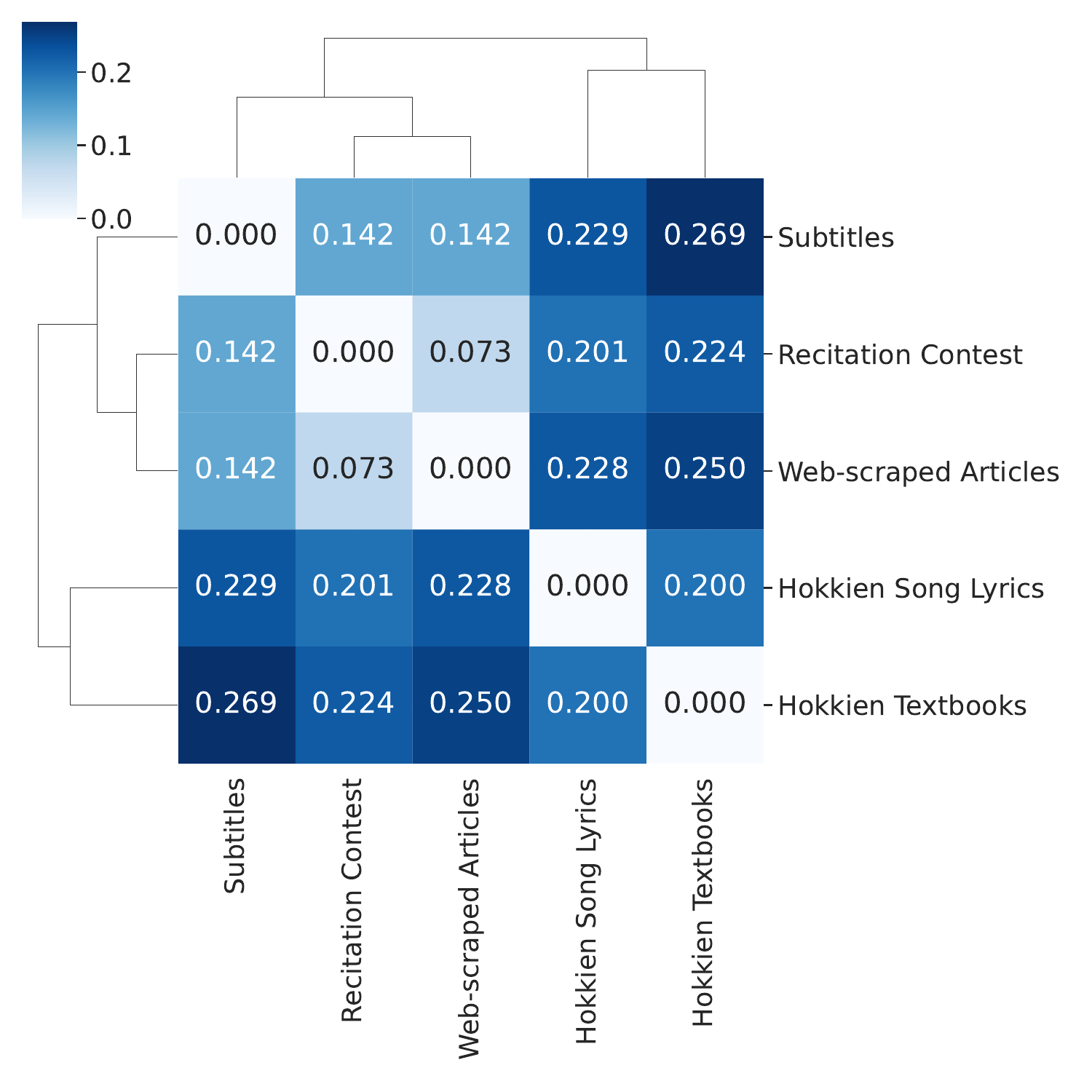} 
\caption{Continued pre-training corpora JSD with dendrogram}
\label{CP_HAN_JSD}
\end{center}
\end{figure}

We use Jensen-Shannon divergence (JSD) as a metric to assess the domain similarity of the monolingual HAN corpora. Due to the absence of open-sourced word segmentation tools for HAN, calculating JSD presents a notable challenge. However, the HAN writing system's resemblance to Traditional Chinese characters allows us to utilize the CKIP Traditional Chinese segmentation tool\footnote{\href{https://github.com/ckiplab/ckiptagger}{https://github.com/ckiplab/ckiptagger}} for processing the HAN corpora. We then computed the JSD and analyzed the domain similarity of our corpora.

\autoref{CP_HAN_JSD} demonstrates that the recitation contest and web-scraped article corpora are more closely aligned. This similarity can be attributed to the fact that both corpora primarily consist of well-structured prose articles, which often explore topics related to local culture, customs and traditions. The content in Hokkien textbooks is composed of verses resembling children's rhymes, making it more akin to Hokkien song lyrics. Subtitles exhibit a distinctive divergence as they primarily feature colloquial sentence structures, setting them apart from the other two groups.

\subsection{JSD of Fine-tuning Parallel Data}\label{sec:appendix_ft_data_jsd}

\begin{figure}[htp!]
\begin{center}
\includegraphics[scale=0.6]{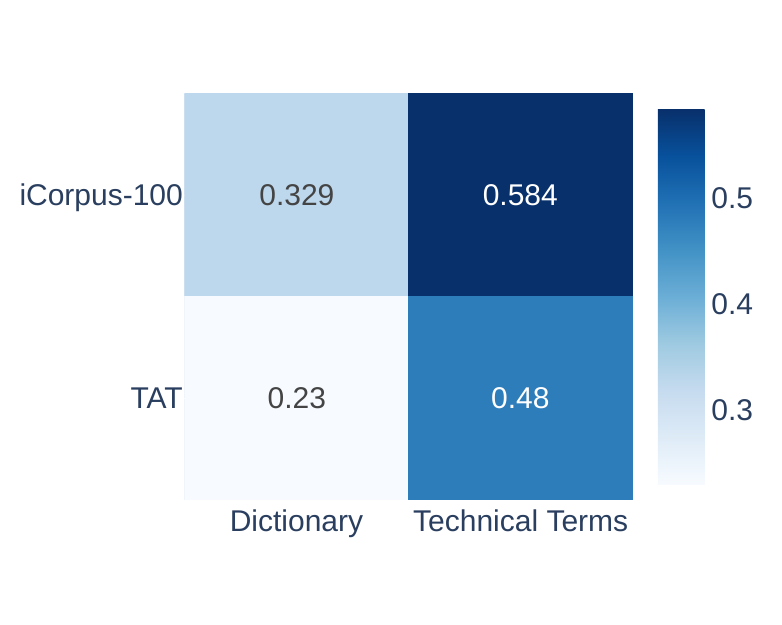} 
\caption{Fine-tuning data JSD}
\label{FT_HAN_JSD}
\end{center}
\end{figure}

In the parallel datasets, the CKIP word segmentation tool is employed to process the Traditional Chinese texts. Consequently, we calculate JSD similarity scores based on the Traditional Chinese portion of the texts. \autoref{FT_HAN_JSD} illustrates the similarity between the training set, which includes both dictionary and technical terms, and the test set, encompassing iCorpus-100 and TAT. We observe that the technical terms subset diverges significantly from the others, because it consists solely of terminology, making it less similar to more general sentences. Moreover, there are significant domain differences between the training and test sets, indicating that achieving a high performance on the translation with this test set presents a considerable challenge.

\section{Evaluating GPT-4's Translation Performance on Hokkien}
\label{sec:appendix-gpt4-translation-performance}

\revised{To evaluate GPT-4's translation performance on Hokkien, we conducted experiments prompting it to translate between Hokkien and both ZH and English. When prompted to translate into HAN, GPT-4 predominantly generated output in ZH, with a limited mixture of HAN and Cantonese words. This finding led us to conclude that this approach is not suitable for assessing back-translation accuracy, as it primarily evaluates GPT-4's translation capabilities in ZH, rather than Hokkien. Additionally, when prompted to translate into POJ, GPT-4's output was completely incomprehensible.}

\revised{Consequently, we only present the results where the target language is ZH or English in \autoref{tab:gpt_ablation_exp}. When translating from HAN, GPT-4 outperforms our best model by $11.15$ and $15.95$ points on the GPT-4 score for HAN-ZH and HAN-EN translation tasks, respectively. Apart from its significantly larger model size, GPT-4's superior performance might be attributed to the similarity writing system between ZH and HAN\footnote{When directly comparing HAN and ZH sentences in iCorpus-100 dataset, we obtain a BLEU score of $45.89$ and chrF++ of $44.91$.}, allowing it to process HAN as a noisy version of ZH and leverage its knowledge of ZH. In contrast, when the source language is POJ, GPT-4 struggles to produce meaningful translations, performing worse than our model with GPT-4 scores of $37.65$ and $20.4$ points for POJ-ZH and POJ-EN, respectively. This emphasizes the need for a specialized large language model designed for Hokkien, which this research aims to address.}

\begin{table}[ht]
\centering
\resizebox{\linewidth}{!}{%
\begin{tabular}{l||cccc} 
\toprule
\multicolumn{1}{c||}{\textbf{Model}} & \textbf{BLEU} & \textbf{chrF++} & \begin{tabular}[c]{@{}c@{}}\textbf{GPT-4}\\\textbf{Score}\end{tabular} & \begin{tabular}[c]{@{}c@{}}\textbf{GPT-4}\\\textbf{Accuracy}\end{tabular} \\ 
\toprule
 & \multicolumn{4}{c}{\textbf{HAN}-ZH} \\ 
\midrule
Our Best & 32.60 & 34.91 & 75.35 & 63 \\
GPT-4 & \uline{59.47} & \uline{59.07} & \uline{86.50} & \uline{81} \\ 
\midrule
 & \multicolumn{4}{c}{\textbf{HAN}-EN} \\ 
\midrule
Our Best & 18.45 & 44.92 & 66.75 & 38 \\
GPT-4 & \uline{33.74} & \uline{58.36} & \uline{82.70} & \uline{79} \\ 
\midrule
 & \multicolumn{4}{c}{\textbf{POJ}-ZH} \\ 
\midrule
Our Best & \uline{30.08} & \uline{31.36} & \uline{43.10} & \uline{12} \\
GPT-4 & 12.19 & 16.65 & 5.45 & 1 \\ 
\midrule
 & \multicolumn{4}{c}{\textbf{POJ}-EN} \\ 
\midrule
Our Best & \uline{7.83} & \uline{30.53} & \uline{24.20} & \uline{3} \\
GPT-4 & 1.78 & 22.69 & 3.80 & 0 \\
\bottomrule
\end{tabular}
}
\caption{The translation performance of GPT-4 on iCorpus-100 datasets. \uline{underline} = the best results for the respective metric.}
\label{tab:gpt_ablation_exp}
\end{table}

\end{CJK*}
\end{document}